 \documentclass[pmlr,twocolumn,10pt]{jmlr} 

\usepackage{booktabs}
\usepackage{siunitx}

\theorembodyfont{\upshape}
\theoremheaderfont{\scshape}
\theorempostheader{:}
\theoremsep{\newline}

\jmlrvolume{LEAVE UNSET}
\jmlryear{2023}
\jmlrsubmitted{LEAVE UNSET}
\jmlrpublished{LEAVE UNSET}
\jmlrworkshop{Machine Learning for Health (ML4H) 2023} 

\title[A Multimodal Dataset of 21,412 Recorded Nights for Sleep and Respiratory Research]{A Multimodal Dataset of 21,412 Recorded Nights for Sleep and Respiratory Research}

\author{
  \Name{Alon Diament} \Email{alon@pheno.ai}\\
  \addr Pheno.AI, Tel-Aviv, Israel\\
  \Name{Maria Gorodetski} \Email{maria@pheno.ai}\\
  \addr Pheno.AI, Tel-Aviv, Israel\\
  \Name{Adam Jankelow} \Email{adam@pheno.ai}\\
  \addr Pheno.AI, Tel-Aviv, Israel\\
  \Name{Ayya Keshet} \Email{ayya.keshet@weizmann.ac.il}\\
  \addr{Weizmann Institute of Science, Rehovot, Israel}\\
  \Name{Tal Shor} \Email{tal@pheno.ai}\\
  \addr Pheno.AI, Tel-Aviv, Israel\\
  \Name{Daphna Weissglas-Volkov} \Email{daphna@pheno.ai}\\
  \addr Pheno.AI, Tel-Aviv, Israel\\
  \Name{Hagai Rossman} \Email{hagai@pheno.ai}\\
  \addr Pheno.AI, Tel-Aviv, Israel\\
  \Name{Eran Segal} \Email{eran.segal@weizmann.ac.il}\\
  \addr{Weizmann Institute of Science, Rehovot, Israel}
 }

\begin{document}

\maketitle

\begin{abstract}

This study introduces a novel, rich dataset obtained from home sleep apnea tests using the FDA-approved WatchPAT-300 device, collected from 7,077 participants over 21,412 nights. The dataset comprises three levels of sleep data: raw multi-channel time-series from sensors, annotated sleep events, and computed summary statistics, which include 447 features related to sleep architecture, sleep apnea, and heart rate variability (HRV). We present reference values for Apnea/Hypopnea Index (AHI), sleep efficiency, Wake After Sleep Onset (WASO), and HRV sample entropy, stratified by age and sex. Moreover, we demonstrate that the dataset improves the predictive capability for various health related traits, including body composition, bone density, blood sugar levels and cardiovascular health. These results illustrate the dataset's potential to advance sleep research, personalized healthcare, and machine learning applications in biomedicine.

\end{abstract}

\section{Introduction}
\label{sec:intro}

Obstructive sleep apnea (OSA) is a sleep disorder in which a person’s breathing is interrupted during sleep due to obstruction of the upper airway as a result of relaxation of the throat muscles. This obstruction can lead to pauses in breathing for short periods of time, which can cause loud snoring, reduction in the  blood oxygen levels, stress response, awakening and fragmented sleep. The prevalence of OSA in the general adult population was estimated at 9\%-38\% (for mild OSA), and was higher in men compared with women \citep{senaratna_prevalence_2017}. Prevalence of OSA increases with age, and was estimated at 88\% in men aged 65-69, and 66\% in women \citep{senaratna_prevalence_2017}.

A home sleep apnea test (HSAT) is a non-invasive test used to diagnose sleep apnea. During a HSAT, a patient wears a portable device overnight to monitor their breathing, heart rate, oxygen levels, snoring, and other sleep patterns. The device records this data and it is then analyzed by clinically-validated algorithms that compute apnea-related indices. The data can also be used to study sleep architecture and other aspects of sleep, such as heart rate variability (HRV).

Here we present a detailed analysis of HSAT data from the 10K study, a large-scale longitudinal research \citep{shilo_10_2021} involving over 7,000 adults recruited from 2020 to 2022. We analyze 447 features extracted from these data, that relate to sleep architecture and quality, OSA, and HRV. We provide reference values for key features, and demonstrate that the dataset can enhance the predictive capability for various health related traits, including body composition, bone density, blood sugar levels and cardiovascular health.

\begin{figure}[htbp]
\floatconts
  {fig:channels}
  {\vspace{-2em}
  \caption{Channels and events. 4 raw time series channels, 2 derived time series (body position, heart rate), and annotated events.}}
  {\includegraphics[width=1.0\linewidth]{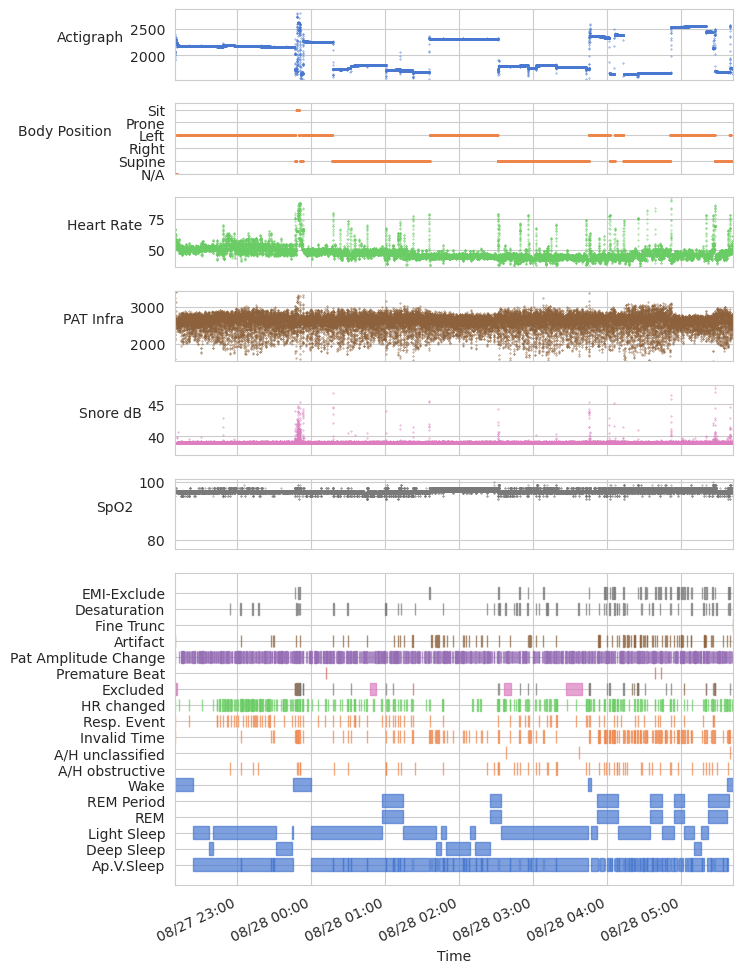}}
\end{figure}
\vspace{-3em}

\section{Materials and Methods}
\label{sec:methods}

\subsection{Study Design}
\label{sec:study}

The 10K study is a longitudinal cohort of deep phenotyping \citep{shilo_10_2021} that is part of the Human Phenotype Project (HPP). The recruited population is of the ages of 40 and 70 years (see the sleep dataset characteristics in \tableref{tab:population}). Follow-up calls and visits are scheduled every year for a total of 25 years. Predefined medical conditions were determined as exclusions. The measured phenotypes in the HPP include a large collection of multi-omics, imaging, mobile sensors, audio recordings, questionnaires and anthropometrics (see also \sectionref{sec:participants}).

The FDA-approved WatchPAT-300 sleep monitor (see \sectionref{sec:device}) is given to participants during the visit to the clinical testing center who then record 3 nights of sleep at home during the following two weeks. The recorded nights are designed to reflect the usual sleeping regimen and typical bedtime.

\subsection{Available Features}
\label{sec:features}

The sleep monitoring dataset is comprised of 3 levels of processing: (1) Time series data for 12 raw, processed, and derived channels (see \sectionref{sec:device}, \figureref{fig:channels}). (2) Annotated events such as sleep stages, classified apnea events, heart rate variability events, and oxygen desaturation events (\figureref{fig:channels}). (3) Summary statistics related to sleep architecture, sleep quality and to OSA, such as the Apnea/Hypopnea Index (AHI), Respiratory Disturbance Index (RDI) and Oxygen Desaturation Index (ODI) indices. Some of the stats are also computed separately for different sleep stages and body positions. Furthermore, we computed features related to HRV (see \sectionref{sec:hrv}). In total these sum up to 447 features. The associations between the various features are shown in \figureref{fig:clustering_sleep,fig:clustering_hrv}.

\section{Results}
\label{sec:results}

\subsection{Reference Values}
\label{sec:reference}

We computed reference values, stratified by age and sex, for four key sleep-derived features (\figureref{fig:ref_ahi_sampen,fig:ref_quality}). These are as follows: AHI, utilized as a diagnostic criterion for OSA \citep{senaratna_prevalence_2017}; sleep efficiency and wake after sleep onset (WASO), two metrics commonly used to diagnose poor sleep quality \citep{ohayon_national_2017}; and sample entropy, a measure of the regularity and complexity within a time series, in this case, the series of R-R intervals between PPG peaks. Lower values signify more regularity and less complexity, and have been associated with pathological conditions and aging \citep{thayer_beyond_2006,goya-esteban_characterization_2008}.

To provide a robust understanding of our sleep-derived features, we averaged measurements over the three nights of monitoring for each participant. This enabled a more accurate representation of each individual's typical sleep behavior. Percentiles for these features were estimated using generalized additive models (GAMs) \citep{serven_pygam_2018}. To evaluate age-related trends, we employed a robust Huber regression model.

Our data show that AHI increases with age (\figureref{fig:ref_ahi_sampen}), corroborating the existing literature on the higher prevalence of OSA among males \citep{senaratna_prevalence_2017}. Intriguingly, the rate of increase in AHI is steeper for females, particularly around the age of 50. For instance, 9\% of females and 24\% of males under 55 have moderate OSA (AHI \(\geq 15\), \citet{senaratna_prevalence_2017}). This rate increases to 30\% for females and 38\% for males above 55. As is commonly observed, sleep efficiency decreases with age, while WASO increases, both trends indicative of deteriorating sleep quality (\figureref{fig:ref_quality}).

We observed that sample entropy diminishes with age for both genders (\figureref{fig:ref_ahi_sampen}), suggesting a decrease in the complexity of HRV features. Sample entropy appears to be higher in females in our data. Interestingly, a marked change in slope is evident for females around the age of 50, potentially indicating hormonal changes or other physiological shifts.

\begin{figure}[htbp]
\floatconts
  {fig:ref_ahi_sampen}
  {\caption{Reference values stratified by age and sex. Dots show data points. Clinical value ranges (green, healthy; red, moderate disease) \citep{senaratna_prevalence_2017}. Dotted lines show percentiles estimated with GAMs. Robust regression equation is given above each plot.}}
  {\includegraphics[width=1\linewidth]{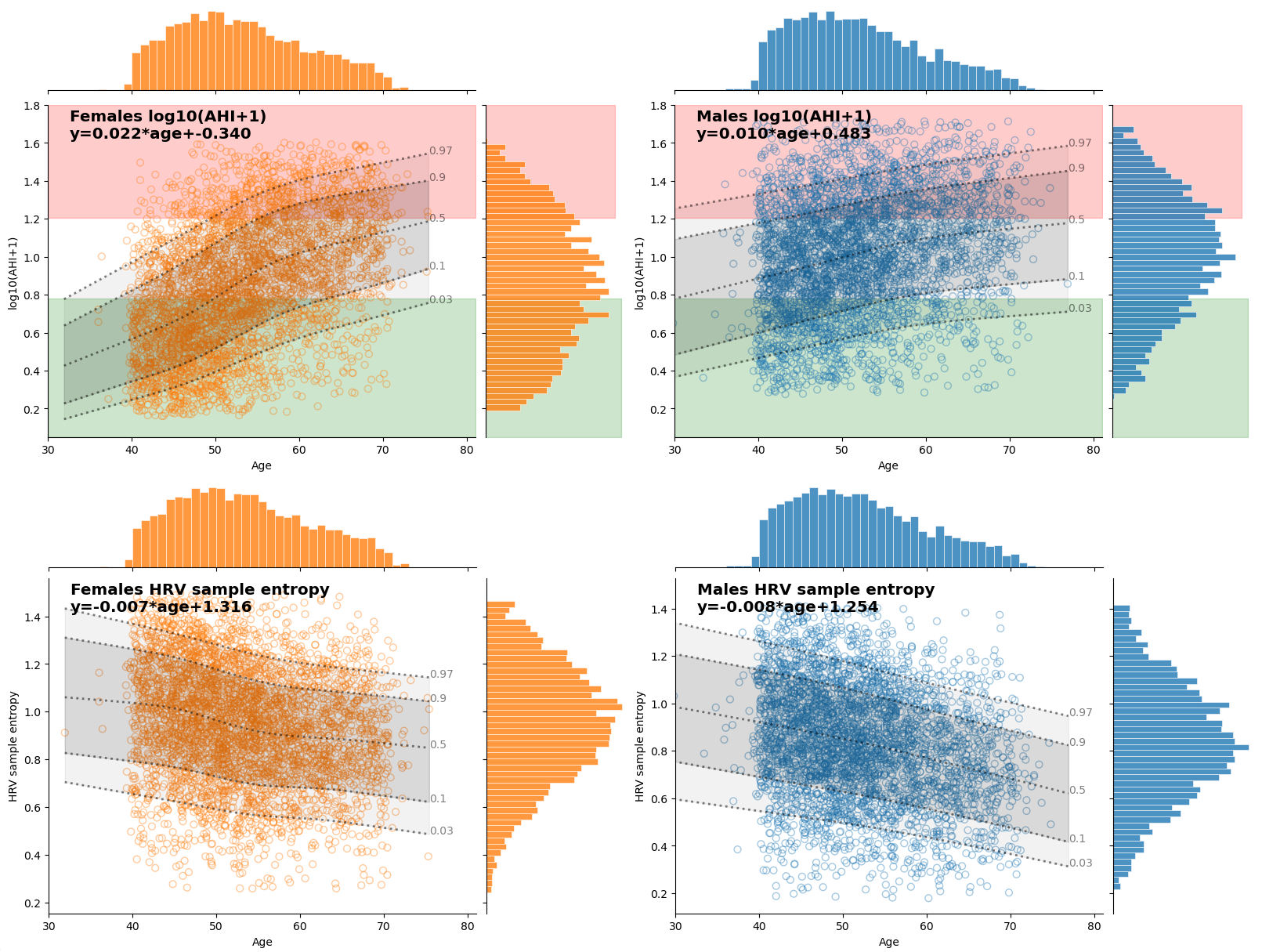}}
\end{figure}
\vspace{-2em}

\subsection{Prediction of Health-Related Traits From Sleep-Derived Features}
\label{sec:prediction}

We employed predictive modeling to demonstrate the clinical utility of our extensive sleep monitoring dataset. We identified three distinct groups of sleep-derived features (\figureref{fig:clustering_sleep}, \figureref{fig:clustering_hrv}): (1) Sleep architecture and quality, which includes metrics like sleep stage duration, sleep efficiency, and sleep fragmentation; (2) Respiratory and OSA, including variables such as AHI, RDI, and measures of oxygen desaturation; and (3) heart rate features, mainly comprising the HRV metrics discussed in \sectionref{sec:hrv}. Although these groupings describe unique dimensions of sleep, it's worth noting that they are not entirely independent. For instance, OSA can affect sleep architecture (see, for example, the total wake time in \figureref{fig:clustering_sleep}).

Prior work highlighted that many sleep-derived features are confounded by BMI, age and sex \citep{van_cauter_metabolic_2008,senaratna_prevalence_2017} (see also \figureref{fig:clustering_sleep}, \figureref{fig:clustering_hrv}). To control for these influences, we established a baseline predictive model that incorporates age, sex, and BMI as its features. Each model was formulated as a Lasso regression to predict a target phenotype using one of the feature groups while including for age, sex, and BMI as covariates. Performance comparisons between the feature-group models and the baseline were conducted using mean absolute error (MAE) (\figureref{fig:prediction}) and correlations (\figureref{fig:spearman}) over cross-validations, and statistical significance was assessed via the paired Wilcoxon sign-rank test.

We incorporated multiple target traits from complementary datasets, specifically from the 10K cohort in the Human Phenotype Project \citep{shilo_10_2021}. These datasets include a wide range of clinically relevant traits, such as imaging-derived features (fundus imaging, DXA scans, liver ultrasound and carotid ultrasound), continuous glucose monitoring (CGM), electrocardiography (ECG), blood pressure measurements, blood tests, and participant responses to physical activity and satisfaction questionnaires. These were carefully chosen to represent a diverse set of healthcare traits.

First, we validated our extracted heart rate features by predicting resting heart rate measured via 10-sec ECG during the participant's visit to the testing center. The predictor based on sleep-derived heart rate features achieved MAE of 5.19 (compared to 7.13 of the baseline predictor) and a correlation of 0.69. This is lower than the error between two ECG measurements from different visits of a participant to the center (6.14), and lower than the MAE of a model that predicts the second visit measurement based on age, sex, BMI, and the first visit (5.87) (\figureref{fig:prediction_visits}).

For a subset of 5,485 participants, CGM was measured concurrently with sleep. The CGM-derived feature, known as eA1C, estimates average glucose levels over the two weeks during which sleep monitoring also took place \citep{keshet_cgmap_2023}. The predictive model using respiratory features significantly improved MAE from a baseline of 0.33 to 0.32 and improved the predictor's correlation from 0.21 to 0.32, in agreement with previous studies on the relation between sleep disordered breathing and diabetes \citep{punjabi_sleep-disordered_2004,west_prevalence_2006}. Although the other feature groups also contributed to this trait-predictor, their impact was more modest.

The physical activity of participants was categorized into walking, moderate, and vigorous activities, based on participant-reported minutes spent daily on each activity. Heart rate features were particularly effective in predicting vigorous activity (decreasing the error from 21.26 at baseline, to 20.16; increasing the correlation from 0.19 to 0.34). This may be due to the more pronounced adaptations in the cardiovascular system commonly associated with exercise, such as improved heart rate variability, lower resting heart rate, and quicker recovery times \citep{routledge_improvements_2010}. However, these heart rate features were less effective for moderate activity and not effective for walking, suggesting that the cardiovascular adaptations might be less apparent or not as relevant for these types of activities. Respiratory features also contributed to these predictive models but to a lesser extent.

Furthermore, heart rate features were significantly associated with ECG QT intervals, diastolic blood pressure, white blood cell count (WBC), visceral and subcutaneous adipose tissue mass (VAT, SAT), bone mineral content and density (BMC, BMD). We included mental well-being, measured through participant-reported life satisfaction, and found that heart rate features were associated with health satisfaction. Conversely, respiratory features associated most closely with sitting blood pressure, non-HDL cholesterol levels, all activity levels, and average vessel width and density extracted from fundus imaging. This aligns with known associations between OSA and cardiovascular health \citep{somers_sleep_2008}.

\begin{figure}[htbp]
\floatconts
  {fig:prediction}
  {\vspace{-1.1em}
  \caption{Improved phenotype prediction using sleep monitoring data. Colors: models based on various feature sets. Baseline: age+sex+BMI predictor. Points: mean absolute error over 4 repeated 5-fold CVs. Bars: CV standard deviation. Asterisks: paired Wilcoxon sign-rank test at 5\% FDR.}}
  {\includegraphics[width=0.75\linewidth]{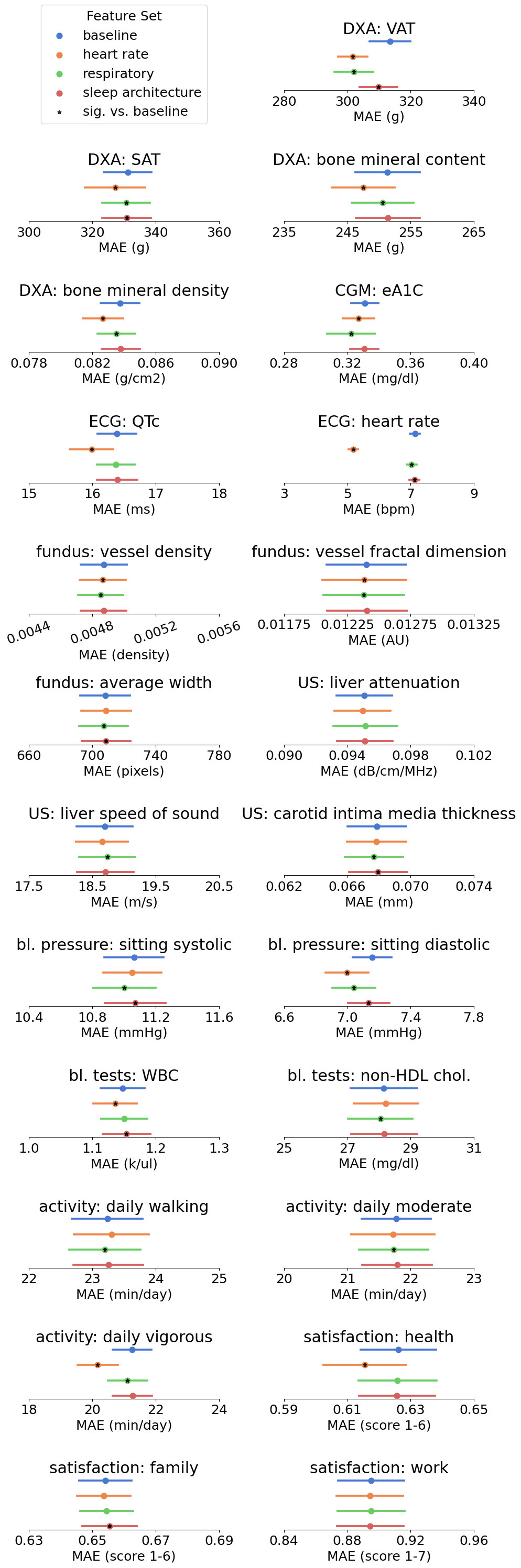}}
\end{figure}

The sleep architecture feature group demonstrated weaker associations with these particular phenotypes. However, it contributed to the predictive models for diastolic blood pressure, VAT and SAT, though to a lesser extent compared to the other feature groups.

We assessed the importance of specific features across various target traits by calculating their selection frequencies with Lasso models (\figureref{fig:selection_quality,fig:selection_respiratory,fig:selection_hrv}). Certain features emerged as 'generalist', frequently selected features across multiple traits; these included age, sex, BMI, average heart rate during wakefulness, and max saturation value. Intriguingly, some features showed a high degree of specificity to particular traits. For instance, the minimum and maximal heart rate recorded during stages of sleep was a frequent predictor of vigorous daily activity. The percent of deep sleep, or slow wave sleep (SWS) was frequently selected to predict VAT, while the total deep sleep time was frequently selected to predict eA1C, in accordance with previous studies on sleep-related metabolic risk factors \citep{van_cauter_metabolic_2008}. These trait-specific features offer promising avenues for future research, and may provide insights into the unique mechanisms underlying each trait.

\section{Discussion}
\label{sec:discussion}

In this study, we introduced a comprehensive dataset of sleep monitoring data, spanning raw time-series, annotated events, and enriched with sleep-derived features. We used it to predict an array of health-related traits including metabolic and cardiovascular phenotypes. The significant associations with clinically relevant phenotypes demonstrate the quality of the data and its utility for future research. It should be noted that these associations should not be interpreted as causal relationships. Importantly, our findings align with existing literature. Additionally, we detected intriguing associations inviting further exploration, such as trait-specific predictive features.

One possible limitation of the provided data is that a recording device could potentially affect some of the measured parameters. The clinically-proven device used in this study has been reported to cause less discomfort, e.g., than respiratory-polygraphy. However, this is an open question regarding sleep monitoring technologies \citep{mueller_sleep_2022,wimaleswaran2018sleeping,metersky_effect_2009}.

We provided a set of descriptive analyses focusing on sleep-derived characteristics of a healthy adult population aged 40 to 70. This age range may be of particular interest for researchers investigating health changes related to aging. Moreover, the dataset holds potential for both cross-sectional and longitudinal research. The current version of the dataset already includes more than 700 participants with multiple monitored visits, and this number is expected to grow considerably in the coming years.

Finally, our dataset opens the door for machine learning applications, particularly deep learning models trained on unstructured time series data from multiple sensors. For example, PPG sensors and accelerometers have been the focus of multiple deep learning models in recent years \citep{jindal_adaptive_2016,biswas_cornet_2019,panwar_pp-net_2020,yuan_self-supervised_2023}.

In summary, the dataset presented here offers a unique and valuable resource for researchers aiming to deepen our understanding of the intricate relationships between sleep and various aspects of human health.

\section{Data Availability}

Data in this paper is part of the Human Phenotype Project (HPP) and is accessible to researchers from universities and other research institutions at the HPP website: \href{http://humanphenotypeproject.org}{humanphenotypeproject.org}.

\bibliography{ml4h_sleep}

\clearpage
\appendix

\renewcommand{\thetable}{S\arabic{table}}
\renewcommand{\thefigure}{S\arabic{figure}}
\setcounter{table}{0} 
\setcounter{figure}{0} 

\section{Supplementary Data}\label{apd:first}

\subsection{Human Participants}
\label{sec:participants}
All participants sign an informed consent form upon arrival to the research site. All identifying details of the participants were removed prior to the computational analysis. The 10K cohort study is conducted according to the principles of the Declaration of Helsinki and was approved by the Institutional Review Board (IRB) of the Weizmann Institute of Science. We analyzed sleep monitoring data of 7,077 recruited individuals, collected between January 2020 and December 2022 (See \tableref{tab:population}).

\begin{table*}[htbp]
\floatconts
  {tab:population}
  {\caption{Characteristics of the study population}}%
  {%
    \begin{tabular}{|l|l|l|l|}
    \hline
    \abovestrut{2.2ex}\bfseries Characteristic & \bfseries Male (n=3,438) & \bfseries Female (n=3,639) & \bfseries All (n=7,077) \\\hline
    \abovestrut{2.2ex}Age & 51.88 (7.85) & 52.94 (7.85) & 52.43 (7.87) \\
    BMI & 26.68 (3.86) & 25.78 (4.46) & 26.21 (4.20) \\
    Recorded nights & 3.06 (1.02) & 2.99 (1.02) & 3.03 (1.02) \\
    \belowstrut{0.2ex}Rec. hours/night & 6.69 (1.38) & 6.87 (1.44) & 6.78 (1.42) \\\hline
    \end{tabular}
  }
\end{table*}

\subsection{Monitoring Device}
\label{sec:device}

The device used for sleep monitoring in this study is the FDA-approved WatchPAT-300 by Itamar Medical. This device contains 5 sensors: An actigraph worn on the wrist; A pulse oximeter and a Peripheral Arterial Tone (PAT) probe worn on the finger; and a microphone and accelerometer worn on the chest for respiratory effort, snoring and body position measurement (RESBP).

The device records 12 raw, processed, and derived channels: actigraph, sleep stage, heart rate, raw heart rate, PAT amplitude, PAT infra, PAT LPF, PAT view, body position, respiratory movement, snore dB, SpO2.

The PAT probe measures changes in the volume of blood vessels by applying a uniform pressure field over the covered finger, preventing venous blood accumulation while measuring changes in the flow within the probe with a photoplethysmogram (PPG) sensor. The measured PAT signal is a physiological signal that reflects changes in the autonomic nervous system caused by respiratory disturbances during sleep.

WatchPAT's algorithm analyzes the PAT signal amplitude, heart rate, and oxygen saturation to identify and classify sleep-related breathing problems. The algorithm provides various indices that allow a diagnosis of sleep apnea, such as the Apnea/Hypopnea Index (AHI). Sleep staging is determined by combining signals from the actigraph and PAT sensors. The accuracy of these algorithms was validated using gold standard polysomnography (PSG) sleep monitoring \citep{yalamanchali_diagnosis_2013,hedner_sleep_2011}.

\subsection{Data Preparation}
\label{sec:preprocessing}

We pre-processed each recording to validate its quality. We detected the end of the night (when the device was removed) to avoid the inclusion of the post-sleep period into analyses. We tested the consistency of the multi-channel time series, events and statistics to validate that they aligned and originated from the same source sample, based on various extracted statistics such as the length of sleep segments. We computed quality scores for each sensor based on the number of exclusion events recorded by the device, where signal quality deteriorated. Signal quality could be low, for example, if one of the sensors was not worn properly, or due to sensor malfunction. In addition, we computed a quality score for HRV features (see \sectionref{sec:hrv}) based on the agreement between the HRV pipeline's estimation of heart rate and the device manufacturer's estimation of heart rate. We excluded recordings that contained less than two hours of sleep, features derived from sources with low signal quality, and features that exceeded predefined minimal and maximal plausible values.

In addition to device-generated statistics, we extracted the following statistics from each night: total wake time after sleep onset (WASO), sleep fragmentation stats (number of stage transitions of each type, time to transition of each type), percent of light / deep / REM sleep, max / min / mean heart rate during sleep stages.

\subsection{Extraction of HRV Features}
\label{sec:hrv}

HRV features can be used to assess the overall health and well-being of an individual \citep{thayer_beyond_2006}. Our sleep monitoring dataset provides a unique opportunity to estimate these features from an extensive recording of several hours at rest. We used NeuroKit2 \citep{makowski_neurokit2_2021} to extract from the raw PAT channel 86 HRV features spanning 5 feature families: time-domain, frequency-domain, nonlinear, complexity / entropy, and fractal dimension. Each of these features is computed on 4 segments of the recordings: entire night, longest non-REM sleep segment, longest REM segment and longest wake segment. Each segment was at least 5 minutes long.

\begin{figure*}[htbp]
\floatconts
  {fig:ref_quality}
  {\caption{Reference values stratified by age and sex. Dots show data points. Clinical value ranges (green, healthy; red, moderate disease) \citep{ohayon_national_2017}. Dotted lines show percentiles estimated with GAMs. Robust regression equation is given above each plot.}}
  {\includegraphics[width=1\linewidth]{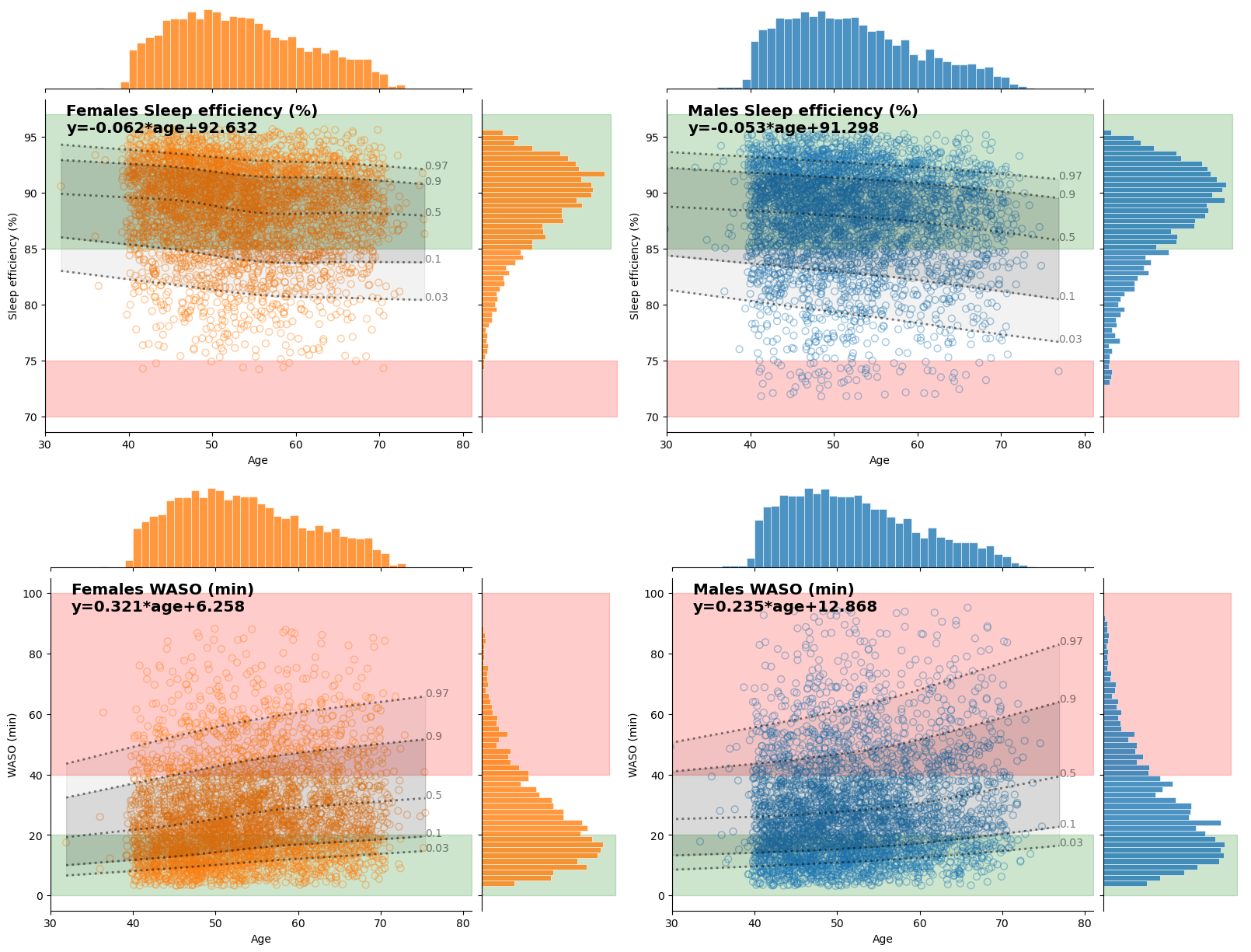}}
\end{figure*}

\begin{figure*}[htbp]
\floatconts
  {fig:clustering_sleep}
  {\caption{Feature correlations: Sleep architecture and respiratory indices. The matrix reports the pairwise Spearman correlation between features. Sleep architecture features are indicated with a plus sign. Age, sex, and BMI are also shown.}}
  {\includegraphics[width=1\linewidth]{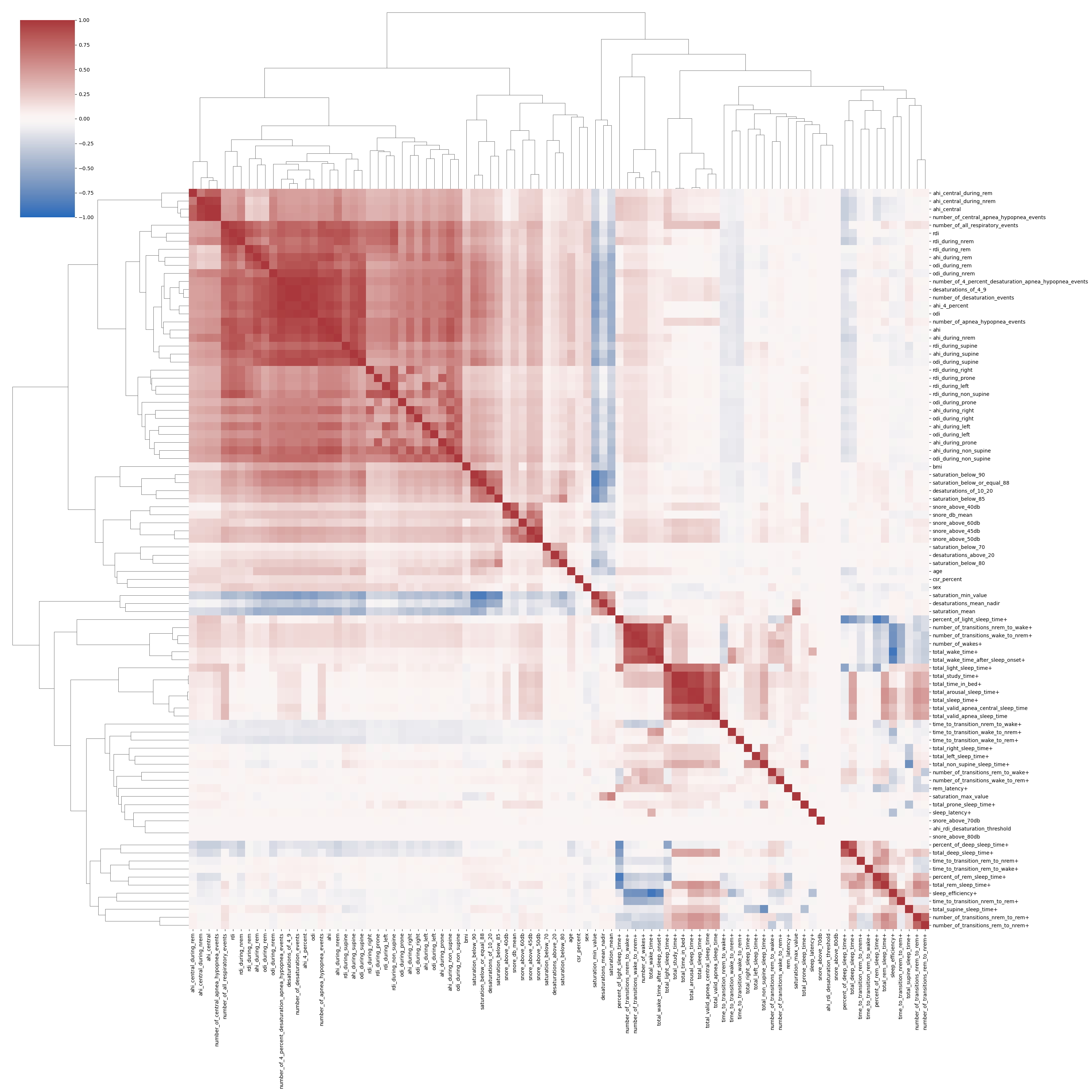}}
\end{figure*}

\begin{figure*}[htbp]
\floatconts
  {fig:clustering_hrv}
  {\caption{Feature correlations: HRV. The matrix reports the pairwise Spearman correlation between features, calculated over the a complete night recording. Age, sex, and BMI are also shown.}}
  {\includegraphics[width=1\linewidth]{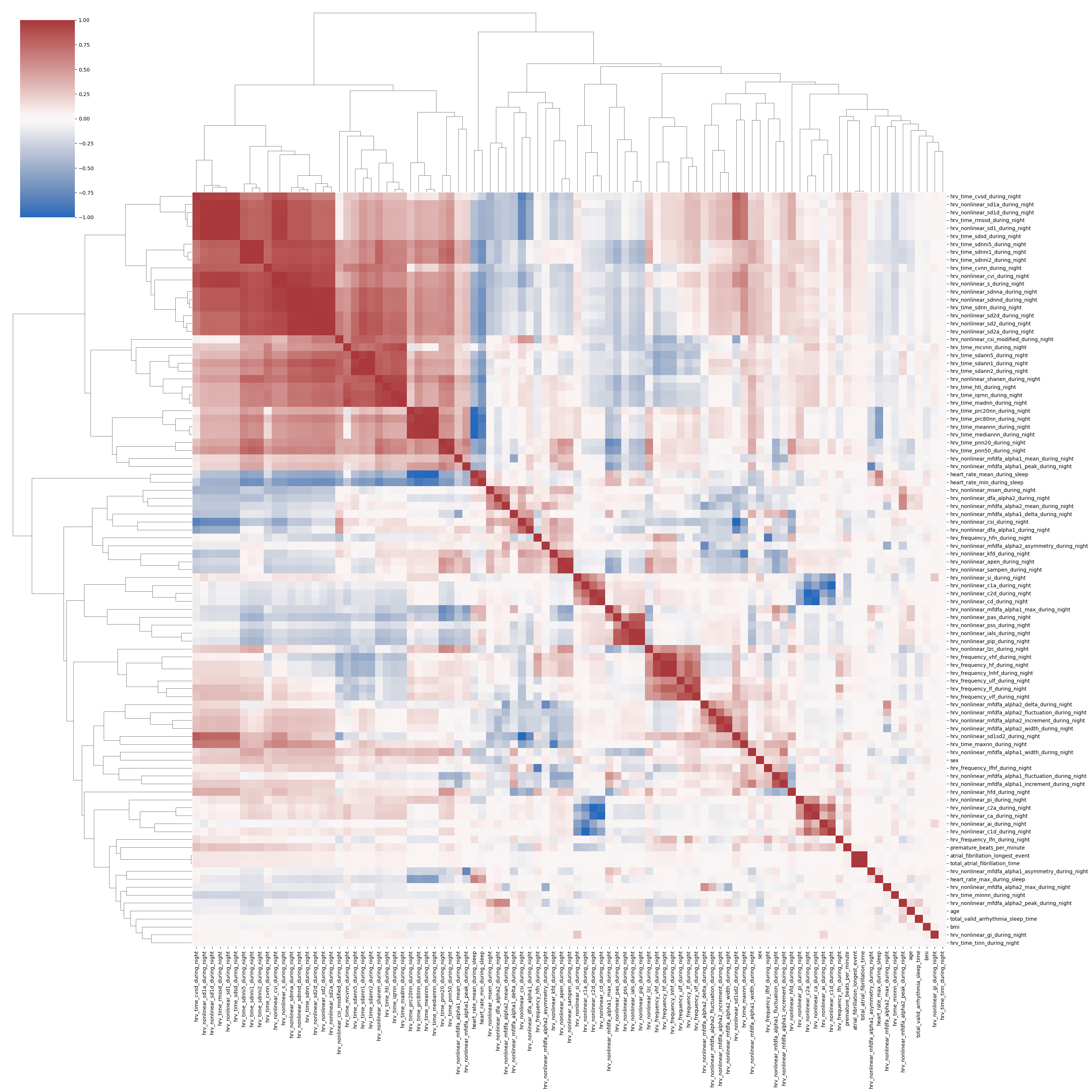}}
\end{figure*}

\begin{figure}[htbp]
\floatconts
  {fig:selection_respiratory}
  {\caption{Feature selection frequency in CV folds: Respiratory features.}}
  {\includegraphics[width=1\linewidth]{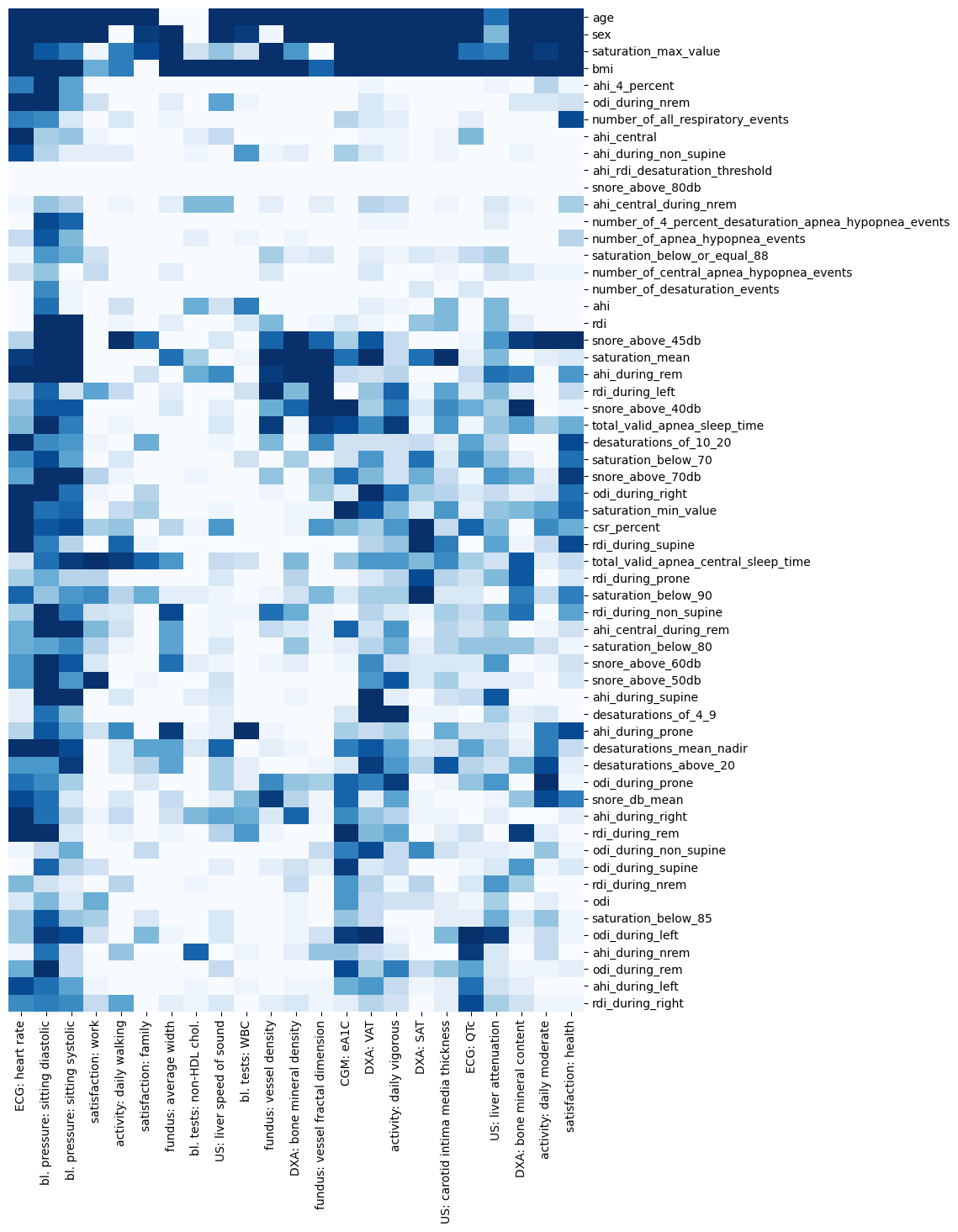}}
\end{figure}

\begin{figure}[htbp]
\floatconts
  {fig:selection_quality}
  {\caption{Feature selection frequency in CV folds: Sleep architecture.}}
  {\includegraphics[width=0.8\linewidth]{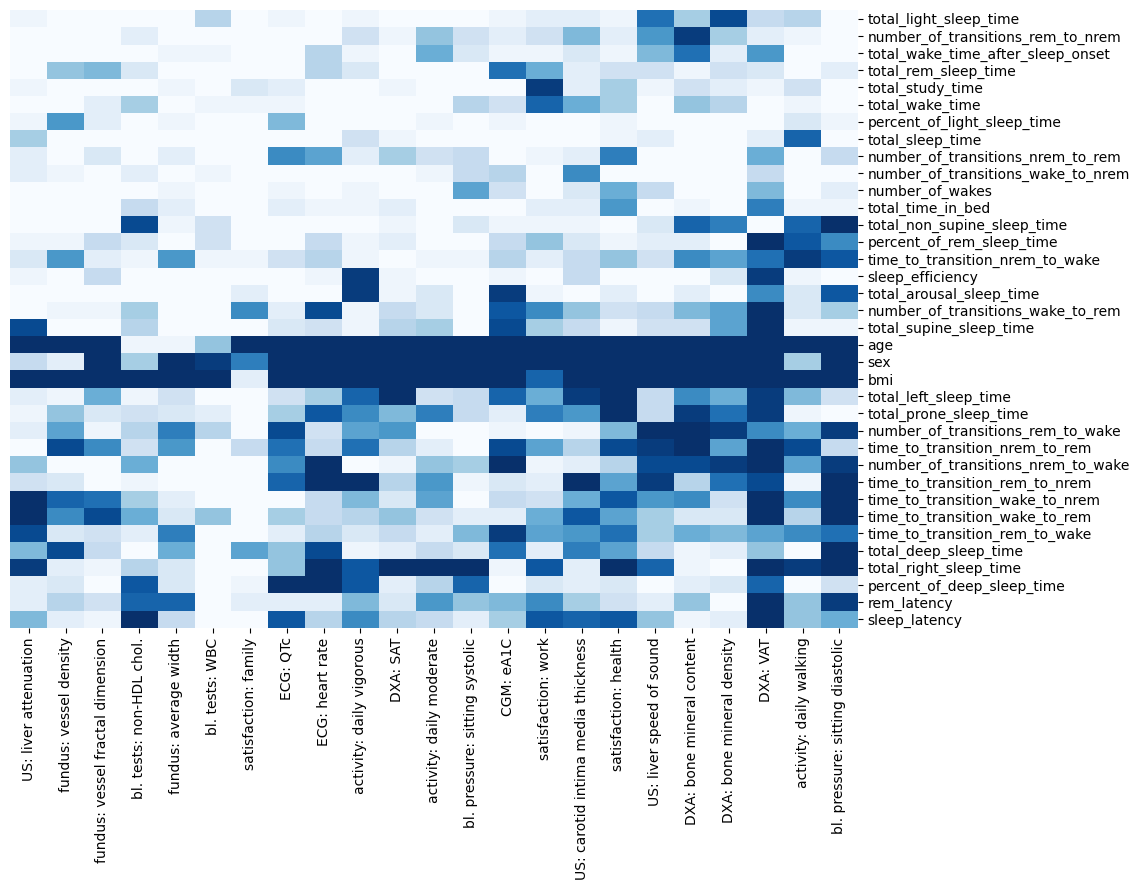}}
\end{figure}

\begin{figure}[htbp]
\floatconts
  {fig:selection_hrv}
  {\caption{Feature selection frequency in CV folds: Heart rate features.}}
  {\includegraphics[width=0.3\linewidth]{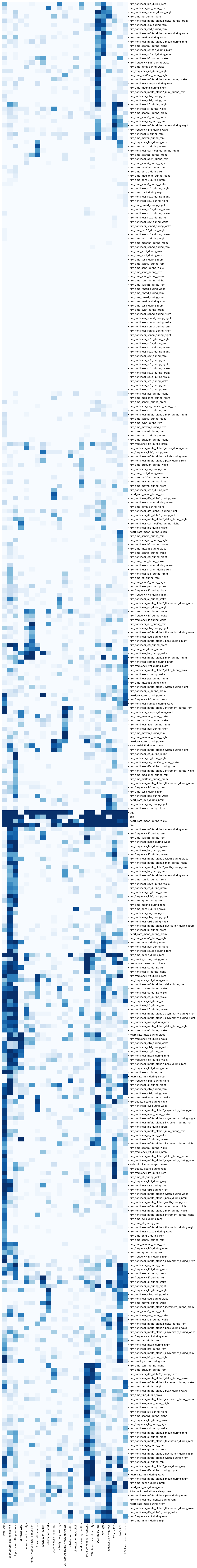}}
\end{figure}

\begin{figure*}[htbp]
\floatconts
  {fig:spearman}
  {\caption{Improved phenotype prediction using sleep monitoring data. Colors: models based on various feature sets. Baseline: age+sex+BMI predictor. Points: Spearman correlation over 4 repeated 5-fold CVs. Bars: CV standard deviation. Asterisks: paired Wilcoxon sign-rank test at 5\% FDR.}}
  {\includegraphics[width=0.9\linewidth]{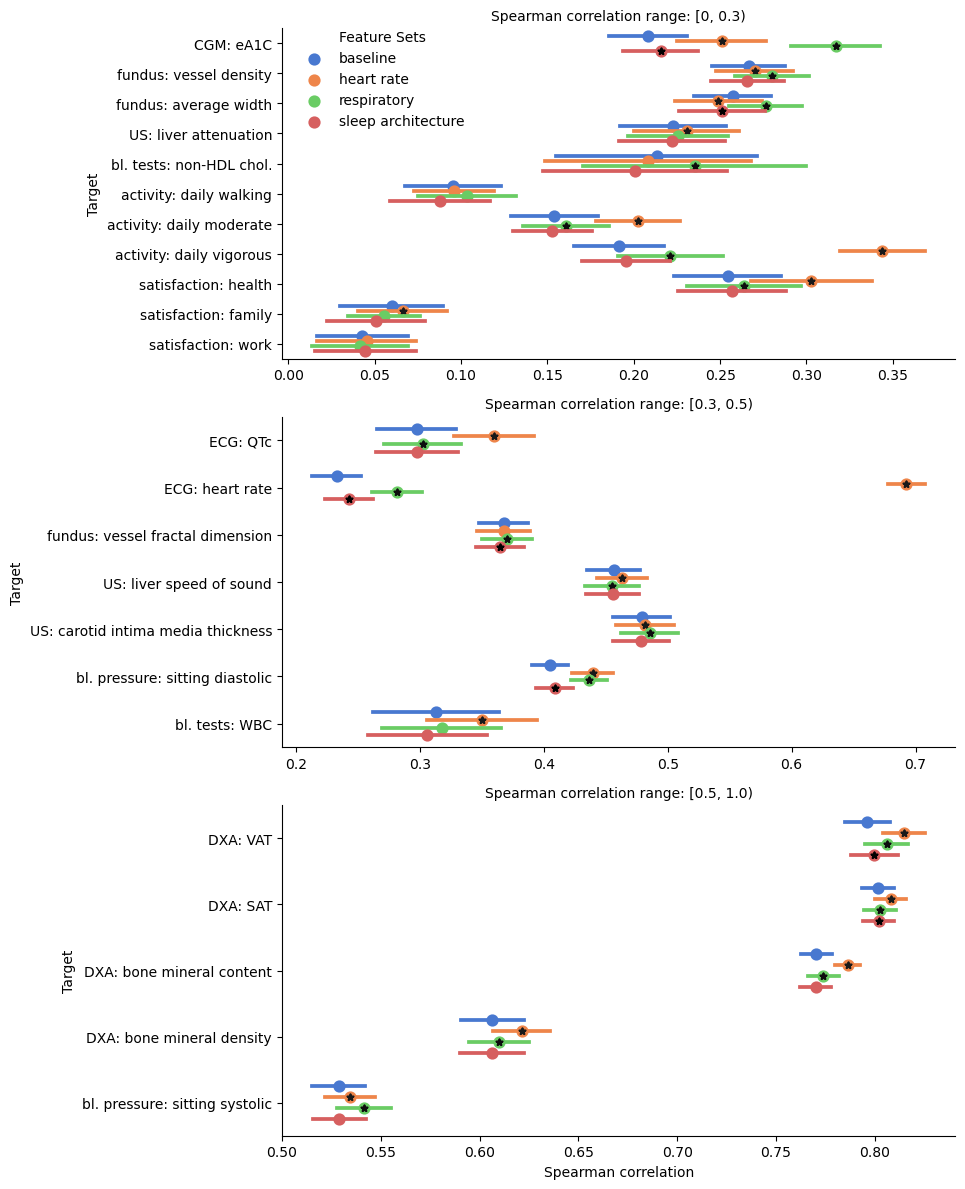}}
\end{figure*}

\begin{figure*}[htbp]
\floatconts
  {fig:prediction_visits}
  {\caption{Supplementary predictions: Using all sleep features, and using data from different visits. Colors: models based on various feature sets. visits-CV: age+sex+BMI+baseline-visit predictor evaluated in CV. visits-MAE: MAE between target variables from two visits. CGM and fundus imaging did not have data from multiple visits. Points: mean over 4 repeated 5-fold CVs. Bars: CV standard deviation.}}
  {\includegraphics[width=1\linewidth]{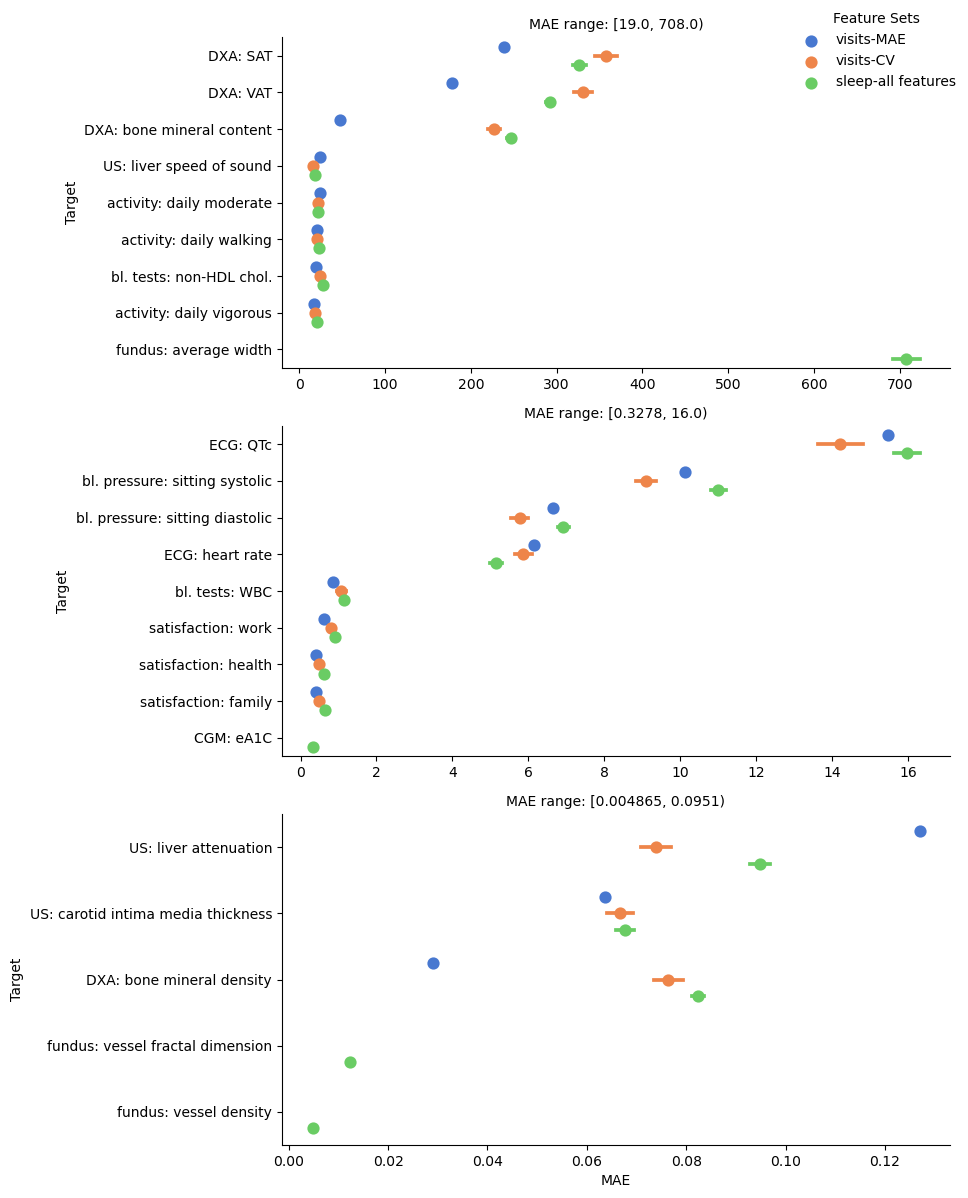}}
\end{figure*}

\end{document}